\documentclass[nonacm,acmlarge]{acmart}

\usepackage{geometry}
\usepackage{float}
\usepackage{indentfirst}
\usepackage[indent]{parskip}
\usepackage{enumitem}
\usepackage{subcaption}

\usepackage{xcolor}
\usepackage{listings}
\usepackage[norndcorners,customcolors]{hf-tikz}

\colorlet{mygray}{black!30}
\colorlet{mygreen}{green!60!blue}
\colorlet{mymauve}{red!60!blue}

\usepackage{algorithm}
\usepackage{algpseudocode}

\newcommand{\gmem}{\lstinline{gmem}}
\newcommand{\smem}{\lstinline{smem}}
\newcommand{\rmem}{\lstinline{rmem}}
\newcommand{\softmax}{\mathrm{softmax}}
\newcommand{\rowmax}{\mathrm{rowmax}}
\newcommand{\rowsum}{\mathrm{rowsum}}

\title[Implementing FlashAttention-2 on NVIDIA Hopper Architecture using the CUTLASS Library]{A Case Study in CUDA Kernel Fusion: Implementing FlashAttention-2 on NVIDIA Hopper Architecture using the CUTLASS Library\vspace{-0.6em}}

\author[Ganesh Bikshandi]{Ganesh Bikshandi$^\dagger$}
\author[Jay Shah]{Jay Shah$^\dagger$}

\authorsaddresses{\textit{Date}: \today. \textit{Email}: \url{research@colfax-intl.com}}

\thanks{\textsuperscript{\dag}Colfax Research. A copy of this paper is available at \url{https://research.colfax-intl.com/nvidia-hopper-flashattention-2/}.}

\begin{abstract}
We provide an optimized implementation of the forward pass of FlashAttention-2, a popular memory-aware scaled dot-product attention algorithm, as a custom fused CUDA kernel targeting NVIDIA Hopper architecture and written using the open-source CUTLASS library. In doing so, we explain the challenges and techniques involved in fusing online-softmax with back-to-back GEMM kernels, utilizing the Hopper-specific Tensor Memory Accelerator (TMA) and Warpgroup Matrix-Multiply-Accumulate (WGMMA) instructions, defining and transforming CUTLASS Layouts and Tensors, overlapping copy and GEMM operations, and choosing optimal tile sizes for the $Q$, $K$ and $V$ attention matrices while balancing the register pressure and shared memory utilization. In head-to-head benchmarks on a single H100 PCIe GPU for some common choices of hyperparameters, we observe 20-50\% higher FLOPs/s over a version of FlashAttention-2 optimized for last-generation NVIDIA Ampere architecture.
\end{abstract}

\begin{document}

\maketitle

\section{Introduction}
For parallel programming on the GPU, one of the most powerful and complex techniques the programmer has at their disposal is \emph{kernel fusion}, which simply refers to the task of combining multiple individual kernels into a new single kernel. The main benefit from kernel fusion comes from reducing the number of reads from and writes to global memory, which is the largest and slowest level of the GPU memory hierarchy. Since contemporary applications are often memory-bound (as opposed to compute-bound) due to growth in GPU compute power outpacing that of memory bandwidth, kernel fusion's importance to breaking through the so-called ``memory wall'' in intensive workloads has only risen with time.\footnote{See \cite{semianalysis} for a polemical take on this, though in a sense NVIDIA's CUTLASS library stands as a rebuttal to that article's main claim.}

Among the most challenging such workloads today involve the training and inference of \emph{large language models} (LLMs). Contemporary LLMs are transformer deep learning models that contain an enormous number of learnable parameters; for example, GPT-3 has about 175 billion parameters \cite{gpt3}. At the heart of the transformer architecture is the attention mechanism \cite{attention}. Attention involves two matrix multiplications and a row-wise softmax operation and is recalled in \S\ref{sec:attention}. Given the centrality of attention to the transformer model, this would appear to be a natural candidate for kernel fusion. It is perhaps surprising then that to our knowledge, the first published attempt to write attention as a fused kernel was only presented in 2022 in the form of the FlashAttention algorithm by Dao et al. \cite{flashattention}, which they present as a ``memory-aware'' version of attention. Subsequently, Dao reworked the algorithm in \cite{flashattention2}, as well as recoding it from the ground up  using NVIDIA's open-source CUTLASS library for high-performance linear algebra \cite{nvidia-cutlass-blog, cutlass}.\footnote{CUTLASS itself was completely rewritten in 2023 for the release of version 3. In particular, the backend core library CuTe is new to version 3.} FlashAttention has since seen widespread adoption and is regarded as the current state-of-the-art \cite{flashadoption,nvidia-records}.

In this paper and the accompanying code, we will take FlashAttention-2 from \cite{flashattention2} as a model example of kernel fusion and consider the engineering challenges involved in implementing it as a CUDA kernel. As in \cite{flashattention2}, we will heavily rely on tools from the CUTLASS library, which greatly simplifies the development of CUDA kernels through the systematic use of abstractions such as Layouts and Tensors. As our interest is primarily didactic, we will constrain ourselves to the forward pass of attention for our study.\footnote{By contrast, the backward pass needed for the backpropagation step during training has a different problem profile as an exercise in kernel fusion. In particular, it involves heavier pressure on the shared memory \cite[\S 2.3.2]{flashattention2}.} Our goal is twofold:
\begin{enumerate}
\item To give an implementation specifically targeting Hopper (SM90) architecture through using Hopper-specific features such as the Tensor Memory Accelerator (TMA) for copying and Warpgroup Matrix-Multiply-Accumulate (WGMMA) instructions for GEMM. In contrast, the implementation in \cite{flashattention2} targets Ampere (SM80) architecture.\footnote{Dao has advertised a comprehensive Hopper-specific implementation as work-in-progress \cite[p. 12]{flashattention2}.} When benchmarked against FLASH-2 from the Dao AI Lab \cite{flash-repo} on a single H100 PCIe GPU, we find 20-50\% higher FLOPs/s in certain representative cases (cf. Figure~\ref{fig:fmha-hopper} in \S\ref{sec:results}).

\item To document and explain some of the challenges and techniques involved that might be generally applicable to problems in kernel fusion, such as the importance of CUTLASS Layouts and transformations thereof. The aim is to complement the discussion in \cite{flashattention,flashattention2} by highlighting some implementation-level details that those papers gloss over.
\end{enumerate}

The accompanying code can be found at \url{https://github.com/ColfaxResearch/cutlass-kernels/tree/master/src/fmha}. We highly encourage the reader to run and examine it alongside reading this paper. 

\section{Standard Attention and Flash (Memory-Aware) Attention} \label{sec:attention}

In this section, we give a rapid review of attention in a transformer model and the FlashAttention-2 algorithm. The input to a transformer model is a batch of tokens of shape $(L = batch\_size, N = seqlen)$. The embedding layer converts this input into a tensor $M$ of shape $(L, N, D = embedding\_dim)$. We then obtain the three $Q$, $K$ and $V$ tensors by multiplying $M$ with three separate trainable weight matrices of square dimension $D$ (as a batched matmul). $Q, K, V$ are then divided along the $D$ mode into $h$ many ``heads'' of head dimension $d = D/h$, so we get these three tensors to be of shape
\[ (L = batch\_size, N = seqlen, h = \#heads, d = headdim). \]

Each head is trained and inferred independently. The choices of $h$ and $d$ depend on the model, but generally $N \gg d$. For example, the \emph{distilbert-base-uncased} model \cite{distilbert} uses $d = 64$, $h = 12$, and $N = 512$ by default.     

Abusing notation, let $Q$, $K$, and $V$ also denote the $N \times d$ matrices associated to a given head. Then the attention output is given by the formula\footnote{One also has optional masking of $S$ and dropout for $P$.}
\[ O = \mathrm{softmax}\left( \frac{1}{\sqrt{d}} Q K^T \right) V = \mathrm{softmax} \left( \frac{1}{\sqrt{d}} S \right)V = P V \]
where $S = \frac{1}{\sqrt{d}} Q K^T$ and $P = \mathrm{softmax}( S)$ are standard variable names for the intermediate expressions. In practice, we replace $S$ by $S - \rowmax(S)$ before taking softmax to avoid overflow with the exponential function; this doesn't change the output of softmax.\footnote{The expression $S - \rowmax(S)$ means we subtract each entry in $S$ by the maximum entry in its respective row. In general, translating a vector by a common value doesn't change the softmax of the vector.} Concatenating over all heads and batches yields the output tensor to be fed into subsequent layers of the model. Observe that the computation of $O$ is independent over the different heads and batches and thus can be executed in parallel. GEMM is also naturally parallelizable along both rows and columns \cite{colfax-gemm}. With this in mind, we have the following naïve (or standard) implementation of attention on the GPU:

\begin{algorithm}[H]
\caption{Standard Attention}
\label{lst:standard}
\begin{algorithmic}[1]
\State Load $Q$ and $K$ by blocks from HBM.
\State Compute $S = (1/\sqrt{d}) Q K^T$ (GEMM-I).
\State Write $S$ to HBM.
\State Read $S$ from HBM.
\State Compute $S = S - \rowmax(S)$.
\State Compute $P = \softmax(S)$.
\State Write $P$ to HBM.
\State Load $P$ and $V$ by blocks from HBM.
\State Compute $O = P V$ (GEMM-II).
\State Write $O$ to HBM.
\end{algorithmic}
\end{algorithm}

Materializing the matrices $S$ and $P$ to HBM (i.e., \gmem) adversely impacts the overall runtime as well as the memory requirement. Indeed, the size of $S$ scales quadratically with the sequence length $N$, which is large (e.g., on the order of 4K or even 32K for state-of-the-art LLM models). Instead, one wants to fuse the individual steps of the attention operation into a single CUDA kernel, thereby bypassing intermediate writes to \gmem. Fusing GEMM with element-wise operations is very straightforward. By comparison, fusing softmax with GEMM is not straightforward as \emph{a priori} softmax involves computing the global maximum and sum along the rows of $S$, while fusion should occur at the threadblock (i.e., CTA) level.

The Fused Multi-Head Attention (FMHA) algorithm in \cite{flashattention2}, taking inspiration from the online-softmax algorithm \cite{online-softmax}, restructures the attention computation to overcome these difficulties and successfully accomplish the fusion.\footnote{Strictly speaking, FlashAttention-2 is an example of an FMHA algorithm, but we will also refer to it as FMHA in this paper for brevity.} Specifically, the FMHA algorithm tiles the matrices $Q$ and $K$ and computes a ``partial'' or ``local'' softmax on the output of GEMM-I, storing the result in local memory (\smem{ }or \rmem). During GEMM-II with tiles of $V$, the partial results are read from local memory, re-scaled with the missing scaling factor and summed back to the result.

\begin{figure}[H]
\begin{subfigure}{.5\textwidth}
  \centering
  \begin{tikzpicture}
  \draw[fill=cyan!60] (0,1.2) rectangle (2,2.2) node[pos=0.5](text=blue) {\tiny $gmem \rightarrow smem$} node at (0.6,2.5) {$Q$};
  \draw[fill=cyan!30] (0,0) rectangle (2,1) node at (2.2,1.1) {$\times$};
  \draw[fill=cyan!60] (2.4,1.2) rectangle (4.4,2.2) node[pos=0.5](text=blue) {\tiny $gmem \rightarrow smem$} node at (3, 2.5) {$K^T$};
  \draw[fill=cyan!30] (2.4,0) rectangle (4.4,1) node at (4.75,1.8) {$=$};
  \draw[fill=cyan!60] (5,1.2) rectangle (6, 2.2) node[pos=0.5](text=blue) {\tiny $rmem$} node at (5.5,2.5) {$S$};
  \end{tikzpicture}
  \caption{GEMM-I, $i = 0$, $j = 0$}
  \label{fig:sfig1}
\end{subfigure}%
\begin{subfigure}{.5\textwidth}
   \centering
  \begin{tikzpicture}
  \draw[fill=cyan!60] (0,1.2) rectangle (1,2.2) node[pos=0.5](text=blue) {\tiny $rmem$} node at (1.0,2.5) {$P=\softmax(S)$};
  \draw[fill=cyan!30] (0,0) rectangle (1,1) node at (2.4,1.1) {$\times$};
  \draw[fill=cyan!30] (1.2,1.2) rectangle (2.2,2.2);
  \draw[fill=cyan!30] (1.2,0) rectangle (2.2,1);
  \draw[fill=cyan!60] (2.6,1.2) rectangle (3.6,2.2) node[pos=0.5](text=blue) {\tiny $gmem \rightarrow$} node at(3.2,1.4) {\tiny $smem$} node at (3.1, 2.4) {$V$};
  \draw[fill=cyan!30] (2.6,0) rectangle (3.6,1);
  \draw[fill=cyan!60] (4,1.2) rectangle (5,2.2) node[pos=0.5](text=blue) {\tiny $rmem$} node at (4.5, 2.4) {$O$} node at (3.8,1.8) {$=$};
  \end{tikzpicture}
  \caption{Softmax+GEMM-II, $i = 0$, $j = 0$}
  \label{fig:sfig2}
\end{subfigure}
\begin{subfigure}{.5\textwidth}
   \centering
 \begin{tikzpicture}
  \draw[fill=cyan!60] (0,1.2) rectangle (2,2.2) node[pos=0.5](text=blue) {\tiny $smem$} node at (0.6,2.5) {$Q$};
  \draw[fill=cyan!30] (0,0) rectangle (2,1) node at (2.2,1.1) {$\times$};
  \draw[fill=cyan!30] (2.4,1.2) rectangle (4.4,2.2) node at (3, 2.5) {$K^T$};
  \draw[fill=cyan!60] (2.4,0) rectangle (4.4,1) node[pos=0.5](text=blue) {\tiny $gmem \rightarrow smem$} node at (4.75,1.8) {$=$};
  \draw[fill=cyan!60] (5,1.2) rectangle (6, 2.2) node[pos=0.5](text=blue) {\tiny $rmem$} node at (5.5,2.5) {$S$};
  \end{tikzpicture}
  \caption{GEMM-I, $i = 0$, $j = 1$}
  \label{fig:sfig3}
\end{subfigure}%
\begin{subfigure}{.5\textwidth}
   \centering
  \begin{tikzpicture}
  \draw[fill=cyan!30] (0,1.2) rectangle (1,2.2) node at (1.0,2.5) {$P=\softmax(S)$};
  \draw[fill=cyan!30] (0,0) rectangle (1,1)  node at (2.4,1.1) {$\times$};
  \draw[fill=cyan!60] (1.2,1.2) rectangle (2.2,2.2) node[pos=0.5](text=blue) {\tiny $rmem$};
  \draw[fill=cyan!30] (1.2,0) rectangle (2.2,1);
  \draw[fill=cyan!30] (2.6,1.2) rectangle (3.6,2.2) node at (3.1, 2.4) {$V$};
  \draw[fill=cyan!60] (2.6,0) rectangle (3.6,1)  node[pos=0.5](text=blue) {\tiny $gmem \rightarrow$} node at(3.2,0.2) {\tiny $smem$} ;
  \draw[fill=red!60] (4.0,1.2) rectangle (5,2.2) node[pos=0.5](text=blue) {\tiny $rmem \rightarrow$} node at(4.6,1.4) {\tiny $gmem$} node at (4.5, 2.4) {$O$} node at (3.8,1.8) {$=$};
  \end{tikzpicture}
  \caption{Softmax+GEMM-II, $i = 0$, $j = 1$}
  \label{fig:sfig3}
\end{subfigure}
\caption{Steps of the FMHA algorithm. \smem{ }and \rmem{ }are used between different stages of computation. Output is written to \gmem{ }in the last stage. Not shown in the figure are the scalings applied to $S$ and $O$ across iterations.}
\label{fig:fmha-steps}
\end{figure}
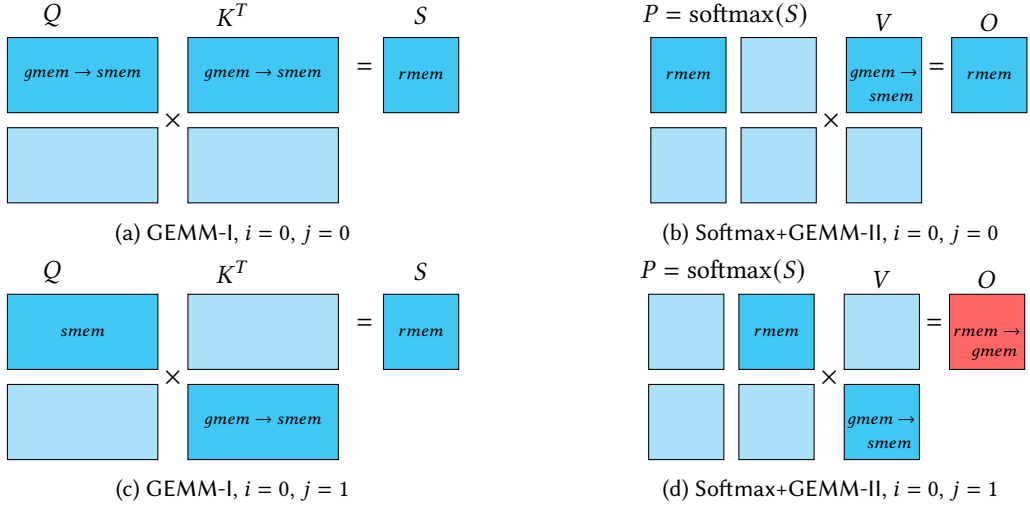

The FMHA algorithm is recalled as Algorithm~\ref{lst:fmha} and illustrated in Figure~\ref{fig:fmha-steps}. We have chosen tile sizes for $Q$ ($bM=\mathrm{QBLK}$) and $K, V$ ($bN=\mathrm{KBLK}$) such that $Q, K, V$ are split into tiles along the row dimension (i.e., M or N dimension), keeping the K-dimension un-tiled ($bK=d$).\footnote{K as the inner dimension for GEMM should not be confused with the matrix $K$.} We have also chosen to display the variant of the algorithm where the first operand for the second GEMM is stored in \rmem\hphantom{ }as opposed to \smem.

\begin{algorithm}
\caption{FlashAttention-2 (FMHA)}
\label{lst:fmha}
\begin{algorithmic}[1]
\For{$i$ \textbf{in} range(tiles of $Q$)} 
    \State Load $bM \times d$ tile $Q_i$ from HBM to SMEM.
    \State Initialize $bM \times d$ accumulator $O_i = (0)$.
    \State Initialize $bM \times 2$ rowmax $m_i = (-\infty)$ and $bM \times 1$ rowsum $\Sigma_i = (0)$.
    \For{$j$ \textbf{in} range(tiles of $K$)}
        \State Load $bN \times d$ tile $K_j$ from HBM to SMEM.
        \State Compute $S_{ij} =  (1/\sqrt{d}) (Q_i K^T_j)$ (SS-GEMM-I).
        \State Update rowmax $m_i = (m_i^{\mathrm{new}}, m_i^{\mathrm{old}})$, tracking rowmax at steps $j$ and $j-1$.
        \State Compute $\widetilde{P}_{ij} = \mathrm{exp}(S_{ij} - m_i^{\mathrm{new}})$.
        \State Update rowsum $\Sigma_i = \exp(m_i^{\mathrm{old}} - m_i^{\mathrm{new}}) \Sigma_i + \rowsum(\widetilde{P}_{ij})$.
        \State Load $bN \times d$ tile $V_j$ from HBM to SMEM.
        \State Compute $O_i = \exp(m_i^{\mathrm{old}} - m_i^{\mathrm{new}}) O_i + \widetilde{P}_{ij} V_j$ (RS-GEMM-II).
    \EndFor
    \State Compute $O_i = (1/\Sigma_i) O_i$.
    \State Write $O_i$ to HBM.
\EndFor
\end{algorithmic}
\end{algorithm}

Accumulators $O_i$, $S_{ij}$, $m_i$, and $\Sigma_i$ are implicitly stored in \rmem. Note that over the outer loop, the algorithm is parallel over threadblocks, while within the outer loop, the algorithm executes within a single threadblock. Therefore, in code the inner loop will appear as the \emph{mainloop} of the computation.

\section{Cutlass/Cute fundamentals, TMA, and WGMMA}

NVIDIA's open-source library CUTLASS and its backend core library CuTe allow one to efficiently write a fused CUDA kernel customized for Hopper (SM90) architecture. In this section, we describe the abstractions and methods from CUTLASS/CuTe that we need to implement Algorithm~\ref{lst:fmha} as a CUDA kernel, including asynchronous \lstinline{copy} and \lstinline{gemm} via Hopper-specific TMA and WGMMA instructions.

\clearpage

\subsection{Layouts and Tensors}

\begin{figure}
\centering
\includegraphics[scale=0.25]{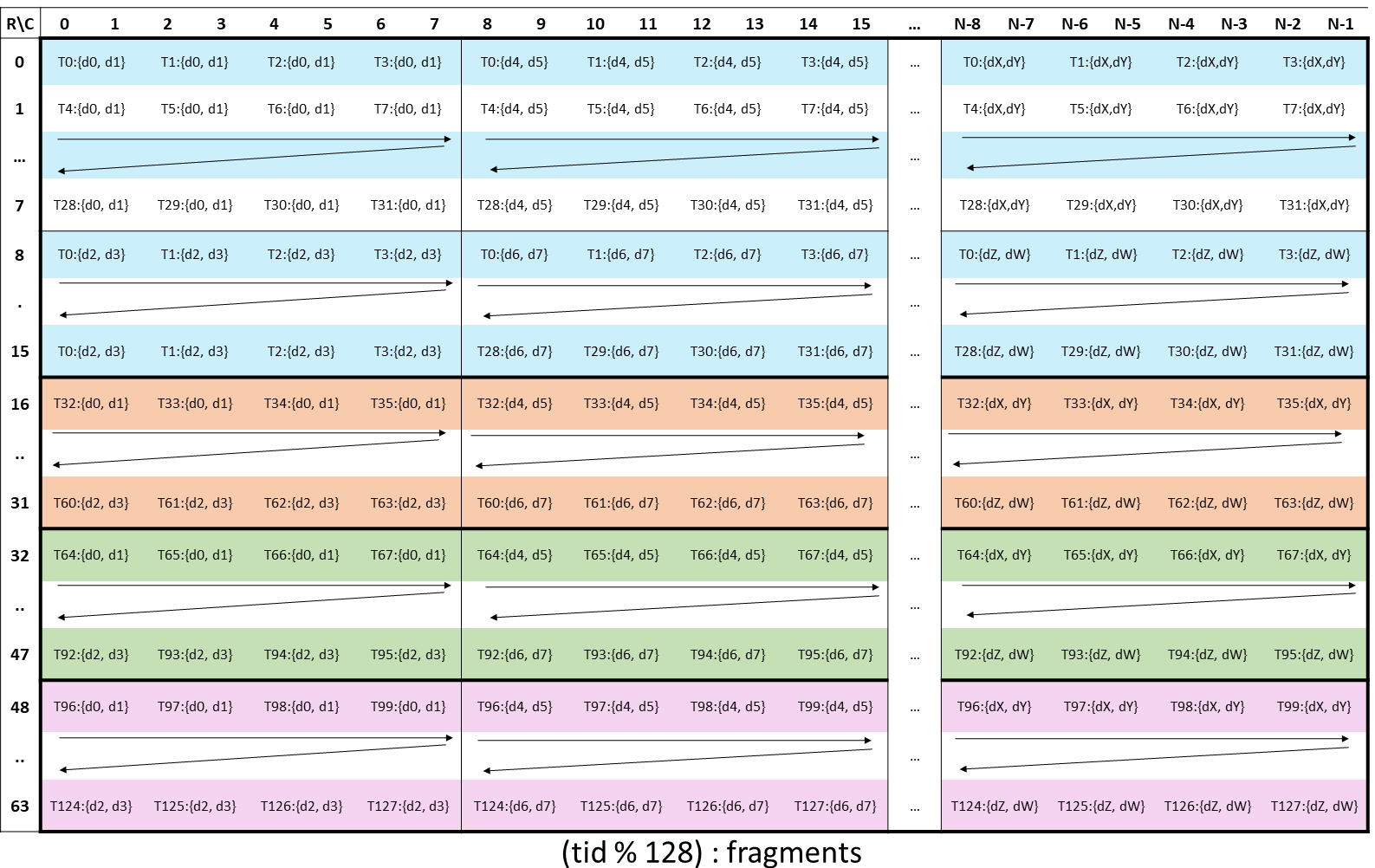}
\caption{Structure of WGMMA accumulator. Take $N$ to be a multiple of 16 for specific choices of tile size, e.g. $N = 64$.}
\label{fig:wgmma}
\end{figure}

The core abstraction of CuTe is the Layout \cite{cutlass-layouts}. Mathematically, a layout $L = \mathbf{n}:\mathbf{d}$ is an object comprised of two integer tuples of common length, the shape $\mathbf{n} = (n_1, ..., n_s)$ and the stride $\mathbf{d} = (d_1, ..., d_s)$,\footnote{These tuples can be also nested but should match in extent, e.g. one could have $L=((2,2),8):((1,2),4)$. The associated layout function is insensitive to parenthesization, but layout operations like reduction along a mode can depend on such.} which determine a multi-linear function
\[ g: [0,n_1) \times \cdots \times [0,n_s) \to \mathbb{N}, \quad (a_1,...,a_s) \mapsto \Sigma_{i=1}^s a_i d_i \]
or equivalently a function of a single variable
$$f = g \circ \iota: [0,\Pi_{i=1}^s n_i) \cong [0,n_1) \times \cdots \times [0,n_s) \to \mathbb{N},$$
where the isomorphism $\iota$ is given by the ``column-major'' traversal. For example, the ``column-major'' layout $L = (4,4):(1,4)$ specifies the identity inclusion
$$f: [0,16) \to \mathbb{N}, \quad i \mapsto i,$$
whereas the ``row-major'' layout $L' = (4,4):(4,1)$ specifies the function sending $[0,16)$ in order to
\[ \{ 0, 4, 8, 12, 1, 5, 9, 13, 2, 6, 10, 14, 3, 7, 11, 15 \}. \]
The use of Layouts is to specify mappings from a logical coordinate (or tuple of coordinates) to a physical coordinate, like an address in HBM or shared memory, or alternatively another logical coordinate, like a mapping of a (thread, value) coordinate to a matrix coordinate for an MMA instruction. For example, the WGMMA instruction with $64 \times 64$ accumulator $C$ has the Layout (in \lstinline{cute/atom/mma_traits_sm90_gmma.hpp}):
\begin{lstlisting}[caption={The $64 \times 64$ accumulator Layout for WGMMA.},label={lst:accumCLayout}]
using CLayout_64x64  = Layout<Shape <Shape <  _4,_8, _4>,Shape < _2,_2,  _8>>,
                              Stride<Stride<_128,_1,_16>,Stride<_64,_8,_512>>>;
\end{lstlisting}

This describes the $(T,V) \mapsto (M,N)$ mapping and corresponds to Figure~\ref{fig:wgmma} \cite[Figure 118 in \S9.7.14]{cuda-ptx}.\footnote{WGMMA is per warpgroup, so involves $128$ threads. To match against $64*64$ entries, one thus has $32$ values per thread. Note how the function associated to \texttt{CLayout\_64x64} restricts to an isomorphism $[0,64*64) \xrightarrow{\cong} [0,64*64)$, and we implicitly have the ``column-major'' isomorphism $[0,64*64) \cong [0,64) \times [0,64)$ for matching the one-dimensional codomain to the two-dimensional logical $(M,N)$ coordinate.}

CuTe Tensors then are constructed from Layouts and pointers into memory, or can be derived from other Tensors (e.g., slicing a Tensor to get a thread-level view). Tensors can be ``owning'' (e.g., in registers) or ``non-owning'' (e.g., a view, in the C++20 sense, of global or shared memory) \cite{cutlass-tensors}.

Though these abstractions may seem complicated at first glance, a working understanding of Layouts and Tensors is essential for developing a fused kernel with CUTLASS. For example, we will see the necessity of ``reshaping'' an accumulator layout (from GEMM-I) to an operand layout (for GEMM-II).

\subsection{TMA Copy}

Hopper introduces the dedicated Tensor Memory Accelerator (TMA) unit for asynchronous copying from \gmem{ }to \smem. TMA is exposed by CUTLASS via \lstinline{make_tma_copy}, which is constructed using the full \gmem{ }Tensor and the target \smem{ }Layout:

\begin{lstlisting}[caption={Constructing the TMA copy object for $Q$}]
auto tileShapeQ = make_shape(bM{}, bK{});
auto smemLayoutQ = tile_to_shape(GMMA::Layout_K_SW128_Atom<MmaA>{}, tileShapeQ);
Layout gmemLayoutQ = make_layout(make_shape(M, K, H, B), make_stride(K * H, 1, K, H * M * K));
Tensor gQ = make_tensor(ptrQ, gmemLayoutQ);
auto tmaQ = make_tma_copy(SM90_TMA_LOAD{}, gQ, smemLayoutQ, tileShapeQ, Int<1>{});
\end{lstlisting}

For optimal performance, we choose the K-major (i.e., row-major) 128-byte swizzling format for \lstinline{smemLayoutQ}; since the atom is premade for us in CuTe, we just need to invoke \lstinline{tile_to_shape} to fit this to the given tile shape.\footnote{Technically, this produces \lstinline{smemLayoutQ} as a ComposedLayout object in CuTe, where one postcomposes the function represented by the Layout with the swizzle function.}  Swizzling in the shared memory is a standard CUDA optimization technique that serves to mitigate bank conflicts \cite{bank-conflicts}. On device, we can then construct thread-level views of \gmem{ }and \smem{ }(in the code, given by the Tensors \lstinline{tQgQ} and \lstinline{tQsQ}) and execute the copy operation using \lstinline{tmaQ}:\footnote{Note though that the TMA programming model is single-threaded - one thread is elected to carry out the copy.}
\begin{lstlisting}[caption={Executing the TMA copy.},label={lst:tmacopy}]
cfk::copy(tQgQ(_, 0), tQsQ(_, 0), tmaLoadQ, tma_load_mbar[0]);
\end{lstlisting}

In Listing~\ref{lst:tmacopy}, our custom \texttt{cfk} method wraps \lstinline{cute::copy} with some synchronization/barrier logic. The other instances of copying from \gmem{ }to \smem{ }are handled similarly.

\subsection{TiledMMA and GEMM}

For optimal performance with Hopper, we want to execute asynchronous WGMMA instructions for matrix multiplication. To facilitate this, CuTe has the MMA atom \cite{cutlass-mma-atom} and the TiledMMA object wrapping it:

\begin{lstlisting}[caption={TiledMMAs for GEMM-I and GEMM-II}]
// USE SS version of GMMA for GEMM-I.
using TiledMma0 = decltype(cute::make_tiled_mma(
      cute::GMMA::ss_op_selector<MmaA, MmaB, MmaC, Shape<bM, bN, bK>>(),
      MmaTileShape{}));
// USE RS version of GMMA for GEMM-II (Default).
using TiledMma1 = decltype(cute::make_tiled_mma(
      cute::GMMA::rs_op_selector<MmaA, MmaB, MmaC, Shape<bM, bK, bN>,
                                 GMMA::Major::K, GMMA::Major::MN>(),
      MmaTileShape{}));
\end{lstlisting}

Here, \lstinline{ss_op_selector} and \lstinline{rs_op_selector} are CuTe helper functions\footnote{Source code in \lstinline{cute/arch/mma_sm90.hpp}.} for selecting appropriate SM90 MMA atoms given the target tile sizes, precision formats (the operand types \lstinline{MmaA} and \lstinline{MmaB} are FP16 while the accumulator type \lstinline{MmaC} is FP32), and choice of whether to put the first operand in \smem{ }(as for GEMM-I) or \rmem{ }(as for GEMM-II). The TiledMMA objects are then used to construct the Tensors that will be  arguments for the \lstinline{gemm} call. For example, we have for GEMM-I:

\begin{lstlisting}[caption={Code for GEMM-I. The \texttt{gemm} call happens within the mainloop.},label={lst:gemmI}]
TiledMma0 tiledMma0;
auto threadMma0 = tiledMma0.get_thread_slice(threadIdx.x);
Tensor tSrQ = threadMma0.partition_fragment_A(sQ);
Tensor tSrK = threadMma0.partition_fragment_B(sK);
Tensor tSrS = partition_fragment_C(tiledMma0, tileShapeS);
// ...
cfk::gemm_bar_wait(tiledMma0, tSrQ, tSrK, tSrS, tma_load_mbar[0]);
\end{lstlisting}

In Listing~\ref{lst:gemmI}, our custom \lstinline{cfk} method wraps \lstinline{cute::gemm} with some synchronization/barrier logic. Similarly, we have for GEMM-II:

\begin{lstlisting}[caption={The \texttt{gemm} call for GEMM-II, within the mainloop.}, label={lst:gemmII}]
cfk::gemm_bar_wait(tiledMma1, convert_type<PrecType, AccumType>(tOrP), tOrV,
                       tOrO, tma_load_mbar[1]);
\end{lstlisting}

\section{Layout Transformations}

Correctly defining the Layouts for the Tensors \texttt{tOrP} and \texttt{tOrV} featuring as operands for GEMM-II in Listing~\ref{lst:gemmII} involves effecting certain transformations of prior Layouts, which we discuss in this section.

\subsection{Taking a Transposed Layout}

By default, a GEMM call in CuTe is in the BLAS NT format, so \lstinline{cute::gemm} takes an $M \times K$-matrix $A$ and a $N \times K$-matrix $B$, and computes $C = A B^T$. Recall that GEMM-I concerns $Q K^T$ while GEMM-II concerns $P V$. Thus, we need to alter the Layout for $V$ (once in \smem){ }so that it can be accepted as the second operand for GEMM-II. Given the \smem{ }Layout for $V$ with shape (bN, bK), we can get to the transposed Layout by a precomposition trick:
\begin{lstlisting}
auto tileShapeV = make_shape(bN{}, bK{});
auto smemLayoutV = tile_to_shape(GMMA::Layout_K_SW128_Atom<MmaB>{}, tileShapeV);
// Layout for Vtranspose. For use in GEMM-II.
auto tileShapeVt = make_shape(bK{}, bN{});
auto smemLayoutVt = composition(smemLayoutV, make_layout(tileShapeVt, GenRowMajor{}));
\end{lstlisting}

Mathematically, given two layouts $L$ and $L'$ with associated functions\footnote{One can always canonically extend the domain of a layout function to all of $\mathbb{N}$ by allowing the last dimension to go to $\infty$.} $f, f': \mathbb{N} \to \mathbb{N}$, one can form the composition $f \circ f': \mathbb{N} \to \mathbb{N}$ and ask whether there is a layout $L''$ such that its associated function $f''$ equals $f \circ f'$; if so, we declare the composition $L \circ L'$ to be given by $L''$. In code, CuTe's \lstinline{composition}{ }function deduces $L''$ for the programmer.\footnote{One should be careful as to whether two given layouts $L$ and $L'$ can actually be composed, which requires satisfying certain divisibility conditions. The \lstinline{composition}{ }function has some static assert checks to rule out impermissible cases, but these aren't comprehensive.} In the case at hand, we are precomposing \lstinline{smemLayoutV} by the layout $(\mathrm{bK}, \mathrm{bN}) : (\mathrm{bN}, 1)$ to take the transposed layout.\footnote{By contrast, if we used \lstinline{GenColMajor} instead of \lstinline{GenRowMajor}, then we would precompose with the identity function and thus do nothing.} Note also that \lstinline{smemLayoutV} involves postcomposing a layout function with a swizzle function, and precomposition by any layout leaves this postcomposition in place.

Given that the \gmem{ }to \smem{ }copy was done with reference to \lstinline{smemLayoutV}, we can then make the transposed Tensor \lstinline{sVt} accessing \smem{ }and correctly define \lstinline{tOrV}:
\begin{lstlisting}
Tensor sVt = make_tensor(make_smem_ptr(shared_storage.smem_v.data()),smemLayoutVt);
// ...
Tensor tOrV = threadMma1.partition_fragment_B(sVt);
\end{lstlisting}
\vspace{-0.8em}

\subsection{Reshaping Accumulator to Operand Layout}

The definition of the Tensor \texttt{tOrP} featuring in GEMM-II involves a custom layout transformation method \texttt{ReshapeTStoTP}. This method takes in the accumulator Tensor \texttt{tSrS} and the Tensor \texttt{tOrS} derived from the TiledMMA object created for GEMM-II, and produces a `reshaped' Layout suitable for defining \texttt{tOrP}:
\begin{lstlisting}
Tensor tOrS = threadMma1.partition_fragment_A(sS);
auto tOrPLayout = ReshapeTStoTP()(tSrS, tOrS);
auto tOrP = make_tensor(tSrS.data(), tOrPLayout);
\end{lstlisting}

The idea is that we need to traverse the accumulator of GEMM-I, held in registers, according to the operand \lstinline{ALayout} selected for TiledMma1. Since we explicitly choose this MMA atom such that operand $A$ is in registers, the relevant \lstinline{ALayout} will look like this (in \lstinline{cute/atom/mma_traits_sm90_gmma.hpp}):
\begin{lstlisting}
// Register source layout for 16-bit value types
using ALayout_64x16 = CLayout_64x16;
// ...
using CLayout_64x16 = Layout<Shape <Shape <  _4,_8, _4>,Shape < _2,_2,  _2>>,
                              Stride<Stride<_128,_1,_16>,Stride<_64,_8,_512>>>;
\end{lstlisting}

Note in particular that the operand \lstinline{ALayout} is also described by Figure~\ref{fig:wgmma} as with the accumulator \lstinline{CLayout}, but now the operand dimensions are fixed to be $64 \times 16$, so each of the 128 threads has 8 values associated to it. To explain further, let's consider the example where tile sizes are all 128 for bM, bN, and bK. Then GEMM-I has accumulator \lstinline{CLayout_64x128}. In this case, printing to console with thread $0$ yields:
\begin{lstlisting}[caption={Tensors and Layouts for tile sizes (bM,bN,bK)=(128,128,128).},label={lst:128layout}]
tSrS: ptr[32b](0x7f25e7fff9e0) o ((_2,_2,_16),_2,_1):((_1,_2,_4),_64,_0)
tOrS: ptr[16b](0x7f25e7fffbe0) o ((_2,_2,_2),_2,_8):((_1,_2,_4),_8,_16)
tOrPLayout: ((_2,_2,_2),_2,_8):((_1,_2,_4),_64,_8)
\end{lstlisting}

For the shapes, the first inner tuple is the value tuple, while the other two coordinates are the column and row coordinates we get from tiling the $128 \times 128$ matrix with the relevant atom shape (either $64 \times 128$ for accumulator or $64 \times 16$ for operand). The Layouts describe a logical to physical mapping where the physical addresses are in \rmem. Observe that for \lstinline{tSrS} and \lstinline{tOrS}, the functions associated to these Layouts, as functions of one variable, are in fact the identity functions.\footnote{However, the extra semantic information encoded by the Layouts (as opposed to their associated functions) is of course important for the reshaping method.} Indeed, we are placing the values for a given thread contiguously in \rmem, and then tiling-to-shape in \emph{column-major} order. On the other hand, the difference between the dimensions $64 \times 16$ and $64 \times 128$ lies in the row dimension. Therefore, we have to traverse \lstinline{tSrS.data()} in a different order than that prescribed by the Layout of \lstinline{tOrS} when defining \lstinline{tOrP}, and thus we change the strides as indicated for \lstinline{tOrPLayout}. By contrast, if we instead had $\mathrm{bM} = 64$, then we wouldn't need to reshape:
\begin{lstlisting}[caption={Tensors and Layouts for tile sizes (bM,bN,bK)=(64,128,128).},label={lst:64layout}]
tSrS: ptr[32b](0x7fed48fffb60) o ((_2,_2,_16),_1,_1):((_1,_2,_4),_0,_0)
tOrS: ptr[16b](0x7fed48fffc60) o ((_2,_2,_2),_1,_8):((_1,_2,_4),_0,_8)
tOrPLayout: ((_2,_2,_2),_1,_8):((_1,_2,_4),_0,_8)
\end{lstlisting}

Finally, note that the reshaping action is decoupled from downcasting the precision format; for the GEMM-II call in Listing~\ref{lst:gemmII}, the \lstinline{convert_type} call on \lstinline{tOrP} will return a new Tensor with \lstinline{tOrPLayout} as its Layout.

\section {Online Softmax and Shuffle Reduction}  \label{sec:online-softmax}

The online-softmax part of the inner loop in Algorithm~\ref{lst:fmha} lying between GEMM-I and GEMM-II involves the row-wise computation of max and sum and altering the $S$ matrix in place. Moreover, we need to successively rescale the matrix $O$ after the first iteration of the loop. In code, we have:
\begin{lstlisting}[caption={online-softmax in the mainloop.}]
if (blockIdxY == 0) { // Compute Online Softmax and NO Output Rescaling.
    onlineSoftmaxAndRescale<true, AccumType>(rowMax, rowSum, tSrS, tOrO, scale); }
else { // Compute Online Softmax and Output Rescaling.
    onlineSoftmaxAndRescale<false, AccumType>(rowMax, rowSum, tSrS, tOrO, scale); }
\end{lstlisting}

After GEMM-I, entries of the matrix $S$ reside in each thread's registers as per Figure~\ref{fig:wgmma}. In particular, values per thread occur over two rows and are traversed in a replicated `Z' pattern; for example, see Listings~\ref{lst:128layout} and \ref{lst:64layout}. In code, this means we need to maintain two values of max when traversing \rmem:
\begin{lstlisting}[caption={Scaling and computing the threadwise rowmax. \lstinline{data} is \lstinline{tSrS.data()}}.]
for (int k = 0; k < NT * size<2>(VT); ++k) {  
    data[n] = FragValType(AccumType(data[n]) * scaleFactor);
    max0 = cutlass::fast_max(max0, AccumType(data[n])); n++;

    data[n] = FragValType(AccumType(data[n]) * scaleFactor);
    max0 = cutlass::fast_max(max0, AccumType(data[n])); n++;

    data[n] = FragValType(AccumType(data[n]) * scaleFactor);
    max1 = cutlass::fast_max(max1, AccumType(data[n])); n++;

    data[n] = FragValType(AccumType(data[n]) * scaleFactor);
    max1 = cutlass::fast_max(max1, AccumType(data[n])); n++;
}
\end{lstlisting}
From Figure~\ref{fig:wgmma} again, we see that a row is partitioned among 4 threads (a \emph{quad}). To assemble these threadwise rowmaxs into the actual rowmax, we could invoke atomic max operations. However, NVIDIA has also provided shuffle instructions to exchange data among threads (from Kepler architecture onwards \cite{shuffle}). We can use these to avoid any atomic operations and thereby avoid the memory access latency inherent to such:
\begin{lstlisting}[caption={Two shuffle reductions for computing the max of two rows.}]
auto max_quad_0 = ShflReduce<4>::run(max0, maxOp);
auto max_quad_1 = ShflReduce<4>::run(max1, maxOp);
mi(rowId) = max_quad_0;
mi(rowId + 1) = max_quad_1;
\end{lstlisting}

The \lstinline{ShflReduce} method is a textbook implementation, but we include it here for completeness:
\begin{lstlisting}[caption=Computing global rowmax using \texttt{shfl} instructions. \lstinline{ShflReduce<4>} computes the global max of a \emph{quad}.,label={lst:shuffle}]
template <typename T> struct MaxOp {
  __device__ inline T operator()(T const &x, T const &y) {
    return x > y ? x : y;
  }
};

template <int THREADS> struct ShflReduce {
  static_assert(THREADS == 32 || THREADS == 16 || THREADS == 8 || THREADS == 4);
  template <typename T, typename Operator>
  static __device__ inline T run(T x, Operator &op) {
    constexpr int OFFSET = THREADS / 2;
    x = op(x, __shfl_xor_sync(uint32_t(-1), x, OFFSET));
    return ShflReduce<OFFSET>::run(x, op);
  }
};

template <> struct ShflReduce<2> {
  template <typename T, typename Operator>
  static __device__ inline T run(T x, Operator &op) {
    x = op(x, __shfl_xor_sync(uint32_t(-1), x, 1));
    return x;
  }
};
\end{lstlisting}
The rowsum computation proceeds similarly.

\section {Overlapping COPY and GEMM} \label{sec:overlap}

One important feature of programming on Hopper GPUs is the ability to overlap asynchronous TMA copy with asynchronous WGMMA instructions in order to hide memory latency and maximize GPU throughput. To accomplish this, CUTLASS recommends the use of software pipelining with multiple buffers for each stage of the pipeline \cite{efficient-gemm}.

Implementing such a scheme is costly from a software engineering point of view, as it would entail large-scale alterations to the structure of the code. Moreover, this would also increase the shared memory requirement by a significant factor. We can instead exploit the structure of the FMHA algorithm, which has two GEMMs occurring within its inner loop. Rather than pipelining for a single GEMM, we issue the loads for GEMM-II with the GEMM-I call and vice-versa. The resulting code is given in Listing~\ref{lst:copy-gemm-overlap}.

\begin{lstlisting}[caption={COPY-GEMM overlapping using TMA+WGMMA},label={lst:copy-gemm-overlap}]  
// Copy first tile of K from GMEM to SMEM.
cfk::copy_nobar(tKgK(_, 0), tKsK(_, 0), tmaLoadK, tma_load_mbar[0]);

for (uint64_t blockIdxY = 0; blockIdxY < nTilesOfK; ++blockIdxY) {
    .....
    // Copy current tile of V from GMEM to SMEM.
    cfk::copy_nobar(tVgV(_, 0), tVsV(_, 0), tmaLoadV, tma_load_mbar[1]);
    clear(tSrS);

    // Issue GEMM-I.
    cfk::gemm_bar_wait(tiledMma0, tSrQ, tSrK, tSrS, tma_load_mbar[0]);
    .....

    // Copy next tile of K from GMEM to SMEM.
    if (blockIdxY != (nTilesOfK - 1)) {
      .....
      cfk::copy_nobar(tKgK(_, 0), tKsK(_, 0), tmaLoadK, tma_load_mbar[0]);
    }
    .....

    // ISSUE GEMM-II with Operand A from RMEM.
    cfk::gemm_bar_wait(tiledMma1, convert_type<PrecType, AccumType>(tOrP), tOrV,
                       tOrO, tma_load_mbar[1]);
}
\end{lstlisting}

\section {Results} \label{sec:results}

\begin{figure}[H]
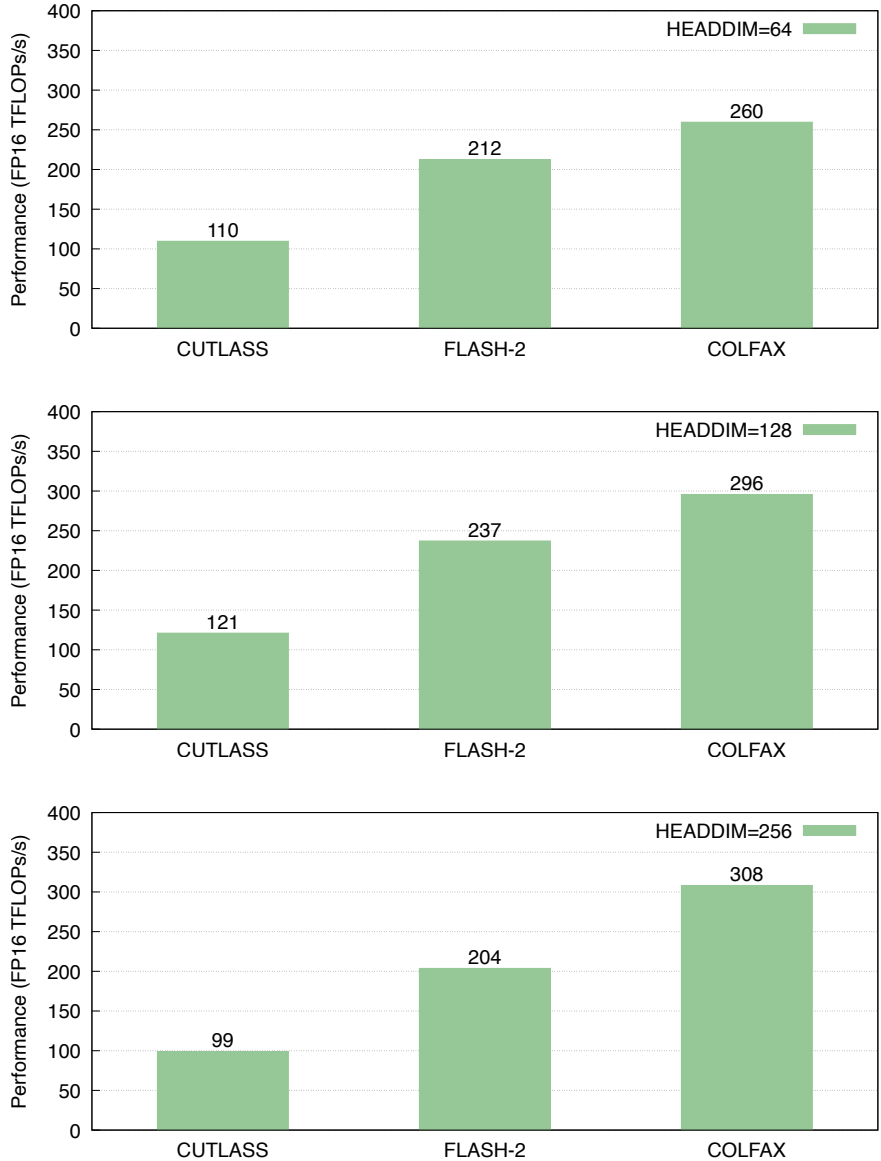

\begin{subfigure}{1.0\textwidth}
  \centering
  \includegraphics[width=0.8\linewidth]{figs/fmha-hopper-fp16-64.png}
  \label{fig:sfig3}
\end{subfigure}
\begin{subfigure}{1.0\textwidth}
  \centering
  \includegraphics[width=0.8\linewidth]{figs/fmha-hopper-fp16-128.png}
  \label{fig:sfig4}
\end{subfigure}
\begin{subfigure}{1.0\textwidth}
  \centering
  \includegraphics[width=0.8\linewidth]{figs/fmha-hopper-fp16-256.png}
  \label{fig:sfig4}
\end{subfigure}
\caption{Performance of FMHA (\emph{forward} pass) on Hopper (H100 PCIe) GPU for $SEQLEN$=$KEYLEN$=4096, $head\_dim$=[64, 128, 256], $num\_heads$=[32, 16, 8], $batch\_size$=4 and FP16 precision with FP32 accumulator. CUTLASS and FLASH-2 versions use SM80 ISA for GEMM and COPY; COLFAX version uses SM90 ISA.}
\label{fig:fmha-hopper}
\end{figure}

We benchmark our implementation of the forward pass of FMHA against two other versions: one shipped as part of CUTLASS 3.3 \cite{cutlass} and FLASH-2 from the Dao AI Lab \cite{flash-repo}. Both of those versions uses SM80 ISA for COPY and GEMM, while we use SM90 ISA (TMA and WGMMA). All of our experiments were conducted on an H100 GPU with PCIe. We summarize the results in Figure~\ref{fig:fmha-hopper}. To interpret these results correctly, the reader should be aware of the following points:
\begin{itemize}[noitemsep]
\item Our COLFAX kernel was compiled with CUTLASS 3.3 and CUDA 12.2.
\item The CUTLASS FMHA kernel is given as example 41 in their codebase and was upstreamed from xFormers \cite{fmha-cutlass}.\footnote{See \url{https://github.com/NVIDIA/cutlass/pull/992}. Note that this dates from before the release of the FlashAttention-2 paper.} We used the FLASH-2 kernel that was part of release 2.3.2.
\item For our program, given the head dimension we experimented with different sizes for $\mathrm{QBLK}$ and $\mathrm{KBLK}$ in the range $(64,128) \times (64,128)$ and chose the best performing one. See Table~\ref{table:tileshapes}.
\item The input matrices had randomly generated values drawn from the Gaussian distribution with mean $0$ and variance $1$, as might be produced after layer normalization.
\item As in \cite[\S 4.1]{flashattention2}, the number of floating point operations was computed in terms of the dominant contributions from the two matmuls, ignoring lower order factors like softmax.\footnote{In contrast, the CUTLASS benchmarking code sums up lower order factors as well when reporting its FLOPs/s computation. We changed this for the common benchmark.}
\item The \texttt{--use\_fast\_math} NVCC compiler flag was used with all three CUDA kernels.
\item Operand types were FP16 and accumulator types were FP32.\footnote{We chose these precision formats for the purposes of reporting benchmarks against the SM80 kernels without changing accuracy across kernels, but we plan to move to lower precision formats in our follow-up work.}
\item We executed the different kernels each with a large number of iterations (iterations=$1000$). Moreover, when rerunning the benchmarking, we observed variations of up to 1 TFLOPs/s.
\end{itemize}
\begin{table}[ht]
\centering
\begin{tabular}{c c c c c}
\hline \hline
HEADDIM &  $(64 \times 64)$ & $(64 \times 128)$ & $(128 \times 64)$ & $(128 \times 128)$ \\
\hline
64 & \textcolor{red}{230.1} & \textcolor{green}{259.5} & 247.9 & 251.4 \\
128 & 292.6 & 289.3 & \textcolor{green}{295.7} & \textcolor{red}{208.7} \\
256 & \textcolor{green}{308.1} & 276.1 & 39.3 & \textcolor{red}{36.7} \\
\hline
\end{tabular}
\caption{Performance with different tile shapes for $\mathrm{QBLK} \times \mathrm{KBLK}$.}
\label{table:tileshapes} 
\end{table}

\vspace{-2em}
We observe that our version achieves speedup close to a factor of 2.5 to 3 over CUTLASS, but only a 20\%-50\% improvement over FLASH-2. 
From our earlier experiments with GEMM \cite{colfax-gemm}, we expected the $128 \times 128$ tile size to deliver the best performance. However, $128 \times 128$ suffers from performance degradation due to register pressure.
We observed register spills with $128 \times 128$ tile size as reported by NVCC.
Additionally, GEMM-II uses both operand A and accumulator C in \rmem.
Sufficient register space is not available to keep them in \rmem{ }at the same time,
due to which the issue of GEMM-II being serialized occurs (as reported by NVCC).

The best performing version uses one warpgroup (128 threads) per CTA, leaving the
register space allowed for another warpgroup wasted.\footnote{The H100 GPU has 64K registers per SM, each of them being 32-bit \cite[\S1.4.1.1]{hopper-tuning}. The maximum number of registers per thread is 255. Using 128 threads utilizes approximately 32K registers.}
In the course of conducting this research, we extended our implementation to use two warpgroups (256 threads) per every tile of operand A of WGMMA. Even though this increases the register space per CTA, the end performance was worse and we chose not to report it in this paper. A better implementation with two warpgroups is work-in-progress.

\section {Future work} \label{sec:conclusion}

The present work originated as part of a larger effort to study CUDA optimization techniques, with a focus on new capabilities afforded by the migration to Hopper architecture. From NVIDIA's H100 datasheet \cite{nvidia-h100-datasheet}, we see that the theoretical maximum TFLOPs/s is 756 for the FP16 Tensor Cores.\footnote{NVIDIA reports 1513 TFLOPs/s for FP16 with sparsity.} As such, we believe that the reservoir of applicable optimizations for the attention problem is far from exhausted. We plan to study at least the following optimizations in future work:
\begin{itemize}[noitemsep]
\item Using two warpgroups (256 threads) per CTA and using a proper warp specialization (WS) scheme;
\item Introducing more pipelining stages into the COPY-GEMM overlapping;
\item Leveraging threadblock clusters and the new distributed shared memory for the $K$ and $V$ matrix COPY.
\end{itemize} 

We also anticipate more implementations of FMHA on next-generation GPU hardware to appear in the near future, and plan to study their methodologies when possible. In particular, we emphasize that SM90 implementations of FMHA (both forward and backward pass) have already appeared as part of LLM libraries used in production. For example, NVIDIA has provided SM90 FMHA kernels as part of its TensorRT-LLM library \cite{tensorrt-llm} that use TMA and warp specialization (WS), though there the kernel source code is not publically available, and OpenAI's Triton also includes an SM90 version in the latest nightly build.\footnote{See https://github.com/openai/triton/pull/2544.}\footnote{We thank Harun Bayraktar and Tri Dao for alerting us to TensorRT-LLM's and Triton's respective SM90 implementations. Since our benchmarking results were conceived in the spirit of an ablation study on the impact of the TMA and WGMMA instructions on performance, we elect to defer a proper comparison with these and other SM90 kernels to future work.} Finally, innovations and advances in GPU architecture beyond Hopper should also provide fruitful ground for revisiting and improving upon the design of FMHA kernels.

\end{document}